\documentclass[10pt,twocolumn,letterpaper]{article}

\usepackage[pagenumbers]{cvpr} 

\makeatletter
\@namedef{ver@everyshi.sty}{}
\makeatother
\usepackage{tikz}

\usepackage{graphicx}
\usepackage{amsmath}
\usepackage{amssymb}
\usepackage{booktabs}
\usepackage{changes}

\usepackage{appendix}
\usepackage{bbm}
\usepackage{multirow}
\usepackage{textcomp}
\usepackage{tabularx}
\usepackage{dsfont}
\usepackage{multirow, makecell}
\usepackage[T1]{fontenc}
\usepackage{algorithm}
\usepackage{algpseudocode}
\usepackage{graphicx}
\newcommand{\bs}{\boldsymbol}
\definecolor{blue}{RGB}{0, 176, 240}
\definecolor{orange}{RGB}{255, 192, 0}
\definecolor{red}{RGB}{255, 0, 0}

\usepackage[pagebackref,breaklinks,colorlinks]{hyperref}

\usepackage[capitalize]{cleveref}
\crefname{section}{Sec.}{Secs.}
\Crefname{section}{Section}{Sections}
\Crefname{table}{Table}{Tables}
\crefname{table}{Tab.}{Tabs.}

\def\cvprPaperID{2366} 
\def\confName{CVPR}
\def\confYear{2023}

\begin{document}

\title{Learning with Noisy Labels over Imbalanced Subpopulations}

\author{\textbf{Mingcai Chen$^{1}$\thanks{Equal contribution. ‡ Work done during an internship at Tencent AI Lab. $\dagger$ Corresponding authors.}\;\footnote[3]{}, Yu Zhao$^{2}$\footnote[1]{}\;, Bing He$^{2}$, Zongbo Han$^{2}$, Bingzhe Wu$^{2}$\footnote[2]{}, Jianhua Yao$^{2}$\footnote[2]{}}\\
$^1$ State Key Laboratory for Novel Software Technology, Nanjing University,\\
$^2$ Tencent AI Lab
}
\maketitle

\begin{abstract}
    Learning with Noisy Labels (LNL) has attracted significant attention from the research community. 
    Many recent LNL methods rely on the assumption that clean samples tend to have ``small loss''. 
    However, this assumption always fails to generalize to some real-world cases with imbalanced subpopulations, i.e., training subpopulations varying in sample size or recognition difficulty.
    Therefore, recent LNL methods face the risk of misclassifying those ``informative'' samples (e.g., hard samples or samples in the tail subpopulations) into noisy samples, leading to poor generalization performance.
    
    To address the above issue, we propose a novel LNL method to simultaneously deal with noisy labels and imbalanced subpopulations. 
    It first leverages sample correlation to estimate samples' clean probabilities for label correction and then utilizes corrected labels for Distributionally Robust Optimization (DRO) to further improve the robustness. 
    Specifically, in contrast to previous works using classification loss as the selection criterion, we introduce a feature-based metric that takes the sample correlation into account for estimating samples' clean probabilities. 
    Then, we refurbish the noisy labels using the estimated clean probabilities and the pseudo-labels from the model's predictions.
    With refurbished labels, we use DRO to train the model to be robust to subpopulation imbalance.
    Extensive experiments on a wide range of benchmarks demonstrate that our technique can consistently improve current state-of-the-art robust learning paradigms against noisy labels, especially when encountering imbalanced subpopulations.
\end{abstract}
\section{Introduction}
\label{sec_intro}
Deep Neural Networks (DNNs) have achieved remarkable progress in various domains, including computer vision \cite{krizhevsky2012imagenet}, health care \cite{healthcare}, natural language processing \cite{DBLP:conf/nips/VaswaniSPUJGKP17}, etc. 
In practice, training datasets may contain non-negligible label noise caused by human annotators' errors.
Therefore, training against noisy labels becomes a critical problem in real-world DNN deployment and has attracted significant attention from the research communities \cite{xiao2015learning,arpit2017closer,DBLP:journals/ml/AngluinL87}.
In recent years, numerous works aim to develop robust learning paradigms to combat label noise \cite{jiang2018mentornet,han2018co,li2020dividemix}.
Among those, estimated clean probabilities are critical for robust training.
For example, Bootstrapping \cite{reed2014training} assigns smaller weights to the loss of possible noisy samples. 
Co-teaching \cite{han2018co} maintains two DNN models, wherein one model is only trained by clean samples selected by another.
Many State-Of-The-Art (SOTA) methods estimate the clean probabilities based on the assumption that correctly labeled samples tend to have ``small loss''.
For example, Dividemix \cite{li2020dividemix} assumes the loss of clean and noisy samples following two Gaussian distributions, while the clean distribution has a smaller mean than the noisy one.
Therefore, it utilizes a two-component Gaussian Mixture Model~(GMM) to model and separate clean and noisy samples.

\begin{figure}[H]
    \centering
    \includegraphics[width=0.9\columnwidth]{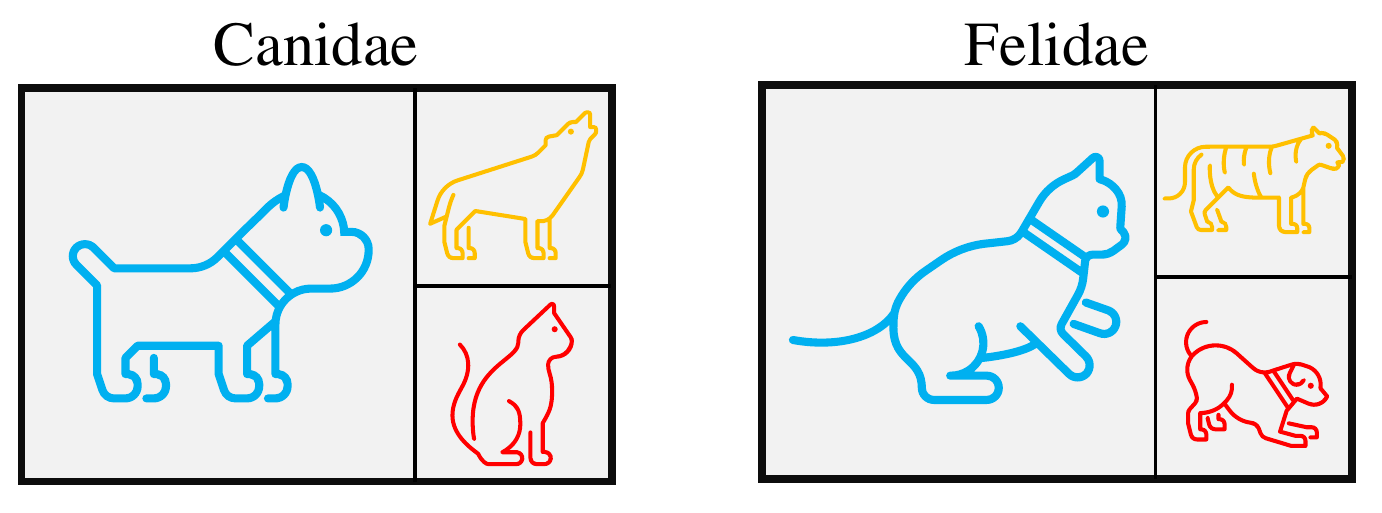} 
    \caption{A demonstration of the learning with noisy labels over imbalanced subpopulations setup. In the Canidae-Felidae classification problem, the \textcolor{blue}{head subpopulations} include dog and cat. The \textcolor{orange}{tail subpopulations} include wolf and tiger. There also exist \textcolor{red}{mislabeled samples}, e.g., cats mislabeled as Canidae or dogs mislabeled as Felidae.
    The size of the images indicates the sample size of the corresponding subpopulation.
    } 
    \label{fig_setting}
\end{figure}

\begin{figure}[H]
    \centering
    \includegraphics[width=\columnwidth]{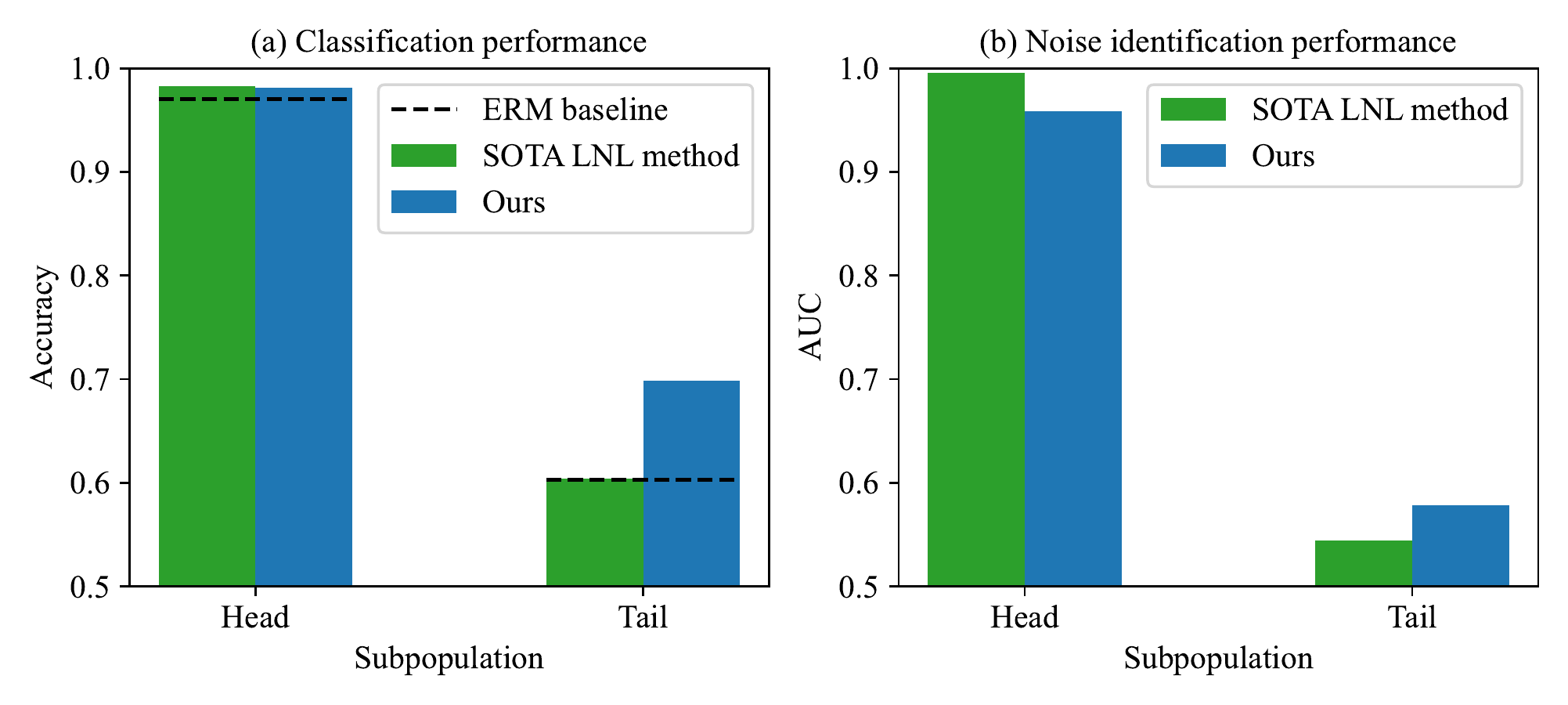} 
    \caption{ (a) Classification accuracy and (b) Noise identification AUC of Empirical Risk Minimization (ERM), SOTA LNL method, and our method under the problem of noisy labels over imbalanced subpopulations. 
    The dataset is the corrupted Waterbirds dataset under 30\% noise rate.
    We use an improved version of DivideMix as a representative LNL method, whose details are introduced in Sec. \ref{sec_experiments}. 
    } 
    \label{fig_confirm}
\end{figure}

Although models trained with these methods can achieve high average performance on the overall population, they may underperform drastically in some subpopulations.\footnote{We use “subpopulation” and “group” interchangeably.}  
The situation could become worse when the training dataset consists of “imbalanced” subpopulations. 
Here, ``imbalanced'' means training subpopulations vary in sample size or recognition difficulty. 
Taking a Canidae-Felidae binary classification problem as an example: the target task is to classify images into two classes, namely Canidae and Felidae, where each class consists of different subpopulations. 
As shown in Fig.~\ref{fig_setting}, there are two subpopulations in each class (the size of the images indicates the sample size of the corresponding subpopulation).
There are also some \textcolor{red}{noisy samples} that adversely affect the training process. 
In such a problem, DNNs can easily overfit on samples in the \textcolor{blue}{head subpopulation} (e.g., dogs) while the \textcolor{orange}{tail samples} (e.g., wolfs) and the \textcolor{red}{noisy samples} (e.g., cats mislabeled into ``Canidae'' class) tend to have large classification loss.
Therefore, previous LNL methods would face the risk of misclassifying tail subpopulations into noisy samples, aggravating the damage of subpopulation imbalance.
We further perform empirical studies on the corrupted Waterbird dataset to quantitatively demonstrate our point. 
Specifically, the representative LNL method (green bar), similar to the ERM baseline (dotted black line), performs much worse on tail subpopulations, as shown in Fig.~\ref{fig_confirm}(a).
We suggest the cause is its bad noise identification performance on tail subpopulations as in Fig.~\ref{fig_confirm}(b).

This paper studies an underexplored novel problem, i.e., learning with noisy labels over imbalanced subpopulations. 
As shown by the above discussions and observations, using the classification loss (e.g., cross-entropy) is inadequate to discriminate between the tail and noisy samples in this problem. 
Therefore, we resort to another way to estimate the label confidence (i.e., the probability of a label being clean) by considering the sample relationship in feature space.  
To this end, we introduce a label confidence estimation criterion named Local Label Consistency (LLC).
Specifically, the LLC of one given sample is obtained by counting the number of samples in the feature-space nearest neighborhood whose labels are consistent with the given sample. 
The design idea behind the LLC metric is that samples from the same class typically come with high feature similarity. 
Utilizing this idea, we employ Gaussian Mixture Models (GMM) for clustering and mapping LLC scores to label confidence.
Next, the label refurbishment framework uses label confidence as a weight to correct the given noisy label.
We further integrate the refurbished labels into Distributionally Robust Optimization (DRO) \cite{DBLP:conf/iclr/SagawaKHL20,DBLP:conf/iclr/ZhangMVBKS21,DBLP:conf/icml/ZhouMMN21}, i.e., constructing the worst-case loss based on estimated label confidence. 
Empirically, we find the proposed refurbish-DRO paradigm can achieve better tail performance.

Through extensive experiments on corrupted Waterbird and CelebA datasets, where subpopulation imbalance and label noise coexist, we show the proposed method consistently improves the performance on tail subpopulations in contrast to current SOTA methods. 
As Fig. \ref{fig_confirm} shows, due to the introduction of LLC-based label confidence estimation and refurbish-DRO paradigm, our method (blue bar) achieves better classification performance and noise identification performance than previous methods (green bar) (details are in Sec. \ref{sec_experiments}). 
Furthermore, our method also improves current state-of-the-art robust training paradigms on standard LNL benchmarks with possible implicit subpopulation imbalance, including CIFAR, Mini-WebVision, and ANIMAL-10N.

Our main contributions are summarized as follows:
\begin{itemize}
    \item We introduce a novel problem, i.e., learning with noisy labels over imbalanced subpopulations, which has been less explored in the community. 
    Moreover, through empirical studies, we demonstrate how previous LNL methods underperform in this new problem.
    \item We propose a general framework for this problem. The basic idea of the framework lies in a simple yet effective strategy that estimates label confidence based on sample relations. The label confidence is then put into our proposed refurbish-DRO paradigm to further improve robustness. 
    \item To verify the effectiveness of our method, we conduct extensive experiments on various datasets, including corrupted datasets with imbalanced subpopulations and noisy labels (Waterbirds and CelebA) and standard LNL benchmark datasets (CIFAR, Mini-WebVision, and ANIMAL-10N). We observe that the proposed method consistently outperforms previous methods. 
\end{itemize}

\section{Related works}
\subsection{Classification-loss-based LNL methods}
Deep models' memorization effect \cite{arpit2017closer} motivated the usage of the ``small-loss'' criterion, which has greatly improved the accuracy of LNL methods.
Lots of them \cite{jiang2018mentornet,han2018co,yu2019does} filter possible noisy samples according to the agreement between the model's predictions and given labels.
For example, DivideMix \cite{li2020dividemix} fits a two-component GMM in every epoch to cluster clean and noisy samples.
Then it utilizes noisy samples by including them for semi-supervised learning.
Similarly, SELFIE \cite{song2019selfie} and RoCL \cite{zhou2021robust} obtain label confidence from cross-entropy-based criterion before using it to correct the noisy labels.
However, we have observed that the classification-loss-based LNL methods obtain sub-optimal performance when training samples form imbalanced subpopulations due to their bad noise identification performance in Sec. \ref{sec_intro}. 

\subsection{Feature-based LNL methods}
Another type of LNL method is built on the assumption that samples in the same class cluster in the feature space. These works are introduced as follows:

\noindent
\textbf{Sample selection} 
Sample selection methods utilize extracted features for the calculation of sample selection criterion, e.g., either directly through the local consistency \cite{DBLP:conf/cvpr/OrtegoAAOM21}, largest connected component on top of a constructed graph \cite{wu2021ngc,wu2020topological}, or principal components from eigendecomposition \cite{DBLP:conf/nips/KimKCCY21}.
However, dropping possible noisy samples leads to the loss of useful information \cite{li2020dividemix}, especially when the subpopulation imbalance exists. 
Our work instead focuses on label confidence estimation for label refurbishment, which better leverages label confidence in a more fine-grained fashion.
What's more, our method retains all samples, including those in tail subpopulations, for training.

\noindent
\textbf{Label refurbishment}
Some label refurbishment methods utilize feature representations to generate new training targets.
D2L \cite{ma2018dimensionality} adopts a label refurbishment framework and obtains optimal label confidence via back-propagating the local intrinsic dimensionality \cite{houle2017local}.
Multi-Objective Interpolation Training \cite{DBLP:conf/cvpr/OrtegoAAOM21} replaces the unreliable labels (i.e., those with little same-class neighbors) with the model’s predictions.
NGC \cite{wu2021ngc} constructs a graph and performs label propagation to obtain pseudo-labels.
On top of this framework, we are motivated to address the new setting where noisy labels and imbalanced subpopulations coexist.
The refurbish-DRO strategy is proposed to encourage models to focus on worst-case performance.

\noindent
\textbf{Regularization}
Iscen et al. \cite{DBLP:journals/corr/abs-2202-02200} proposed an LNL method to directly enforce local consistency using an extra loss term.
Other methods, such as Sup-CL \cite{DBLP:journals/corr/abs-2203-04181}, and NGC \cite{wu2021ngc}, take advantage of contrastive learning. 
Sup-CL selects confident pairs for supervised contrastive learning. 
NGC \cite{wu2021ngc} adds an extra loss term to pull samples in each class's largest connected component together based on a constructed graph.
This paper does not include these regularization strategies for simplicity and generality. However, our method is orthogonal to these regularization methods, which can be combined with our method to probably further improve performance.

\subsection{Subpopulation shift}
Subpopulation shift exists in many real-world data.
For example, in speech recognition, the collected speech is mainly from native speakers.
However, users of the speech recognition system could consist of many non-native speakers, for which the model would yield suboptimal performance \cite{DBLP:conf/icml/HashimotoSNL18}.
Additionally, after suffering from a bad user experience, minority users are likely to stop using the product.
Model retrained on future user data in turn bias toward the majority subpopulation, causing the disparity amplification problem \cite{DBLP:conf/icml/HashimotoSNL18}.

With group information available, some methods mainly use DRO to deal with subpopulation shift \cite{DBLP:conf/iclr/SagawaKHL20,DBLP:conf/iclr/ZhangMVBKS21,DBLP:conf/icml/ZhouMMN21,DBLP:conf/icml/Yao0LZL0F22}.
Some recent works study the domain-oblivious setting, where subpopulation memberships (i.e., subpopulation-level labels) are unknown during training.
\cite{DBLP:conf/icml/LiuHCRKSLF21,DBLP:journals/corr/abs-2007-02561} perform multi-round training.
A preparatory model is normally trained, whose misclassified samples are up-weighted for the model's training in the next round.
On top of this, \cite{DBLP:conf/icml/ZhangSZFR22} 
uses contrastive loss to explicitly encourage the model to learn consistent representation for subpopulations in the second round.
EIIL \cite{DBLP:conf/icml/CreagerJZ21} uses an ERM-trained model to estimate subpopulation, followed by invariant learning \cite{DBLP:journals/corr/abs-1907-02893}.
However, these methods assume that label information is perfect, which could be unrealistic in practice. 
Our work considers a more practical problem setting, i.e., label noise and subpopulation imbalance coexist.

\subsection{Other similar problem settings}
A line of very recent LNL works considers some other realistic settings.
NGC \cite{wu2021ngc} studies the open-world noisy data.
On top of handling the noisy labels, it attempts to detect out-of-distribution samples simultaneously.
Different from this, our setting aims to utilize all in-distribution samples, whether mislabeled or in tail subpopulations, for training instead of excluding out-of-distribution data.
H2E \cite{DBLP:conf/eccv/YiT0LZ22} proposes the noisy long-tailed classification problem.
Our work instead focuses on a more general case, i.e., classes consist of subpopulations with different sizes or learning difficulties.
Besides, our setting assumes the group annotations are unknown (instead of known class imbalance).

\section{Method}
\subsection{Problem statement}
Different from the standard supervised learning, only a noisy training dataset $\mathcal{\tilde{D}}=\{\boldsymbol x_i,\tilde{\boldsymbol{y}}_i\}_{i=1}^{N}$ is available in LNL, where $\boldsymbol x_i$ is the input feature and $\tilde{\bs {y}}_i\in \{0,1\}^C$ denote the one-hot vector of noisy label $\tilde{y}_i\in \{1, \ldots, C\}$. $N$ is the number of training samples, and $C$ is the number of classes.
LNL is to train a robust model which can avoid the impact of noisy labels in the training data and achieve accurate predictions on the unseen clean test dataset. 
The model can be viewed as a composition of a feature extractor $g(\cdot)$ and a linear classifier $f(\cdot)$ working on top of the feature extractor.

In real-world practice, it is a common case that imbalanced subpopulations exist in the dataset.
This paper therefore further assumes that the dataset $\mathcal{\tilde{D}}$ consists of $K$ subpopulations, i.e., $\mathcal{\tilde{D}}= \mathcal{\tilde{D}}_0 \cup \ldots \cup \mathcal{\tilde{D}}_K (K \geq C)$. 
The size or learning difficulty varies across subpopulations, and the subpopulation memberships are unknown during training.
The model is further evaluated on each subpopulation in the clean test set.

\subsection{Local label consistency for label confidence estimation}
\label{sec:lv}
After observing the failure of using classification loss for label confidence estimation when subpopulation imbalance and label noise coexist, we resort to other criteria to tackle this issue.
Considering that the cluster assumption still applies to subpopulations \cite{DBLP:conf/icml/ChapelleCZ06,DBLP:conf/nips/SohoniDAGR20}, we propose to utilize the positive correlation between the label confidence and local agreement in the representation space.

\noindent
\textbf{Local Label Consistency} We design a metric termed Local Label Consistency (LLC) to replace the conventional classification loss for label confidence estimation, which can be calculated as follows. We first obtain penultimate layer features $\bs z_i=g(\bs x_i)$ and calculate the pair-wise cosine similarity $sim(\bs z_i,\bs z_j)=\bs z_i^\mathrm{T}\bs z_j/||\bs z_i||\,||\bs z_j||$.
The k-nearest neighbors of each node (sample) $\mathcal{N}_k(i)$ are selected according to the calculated pair-wise cosine similarities.
Then, the LLC of $i$-th sample is calculated as the proportion of neighbors with the same class label,
\begin{equation}
    \mathrm{LLC}_i=\sum_{j \in \mathcal{N}_k(i)} \mathds{1} [\tilde{\bs y}_i=\tilde{\bs y}_j]
\end{equation}
where $\mathds{1}[\cdot]$ is the indicator function.
It equals 1 if the equation in it holds. Otherwise, it equals 0.

\noindent
\textbf{Label confidence}
After obtaining the LLC scores of samples, we perform unsupervised clustering on these scores to obtain the label confidence, i.e., clean probability: $\mathrm{p}(y=\tilde{y}|\bs x)$, where $y$ is the inaccessible ground-truth label.
Specifically, in every epoch, we use the current model to obtain all samples' LLC scores.
Then, we use a one-dimensional two-component GMM to fit the distribution of these LLC scores. The corresponding parameters are estimated using the Expectation–Maximization (EM) algorithm (We elaborate on the details in the supplementary material.).
After this, The probability of each sample belonging to the GMM component with a bigger mean can be obtained, which is used as the label confidence $w_i$ for $i$-th sample.
The overall pipeline is shown in Alg. \ref{alg}.

\begin{algorithm}
    \caption{Our method} 
    \label{alg}
    \textbf{Input}:
    Noisy dataset $\tilde{\mathcal{D}}=\{(\bs x_i,\tilde{\bs y}_i)\}^{N}_{i=1}$,
    \# epochs for warm up $E_{warmup}$, 
    \# total epochs $E$, 
    ERM training strategy $\text{Train}(\text{dateset},\text{parameters})$,
    robust training strategy with label confidence $\text{RobustTrain}(\text{dateset},\text{label confidence},\text{parameters})$.
    \\
    \textbf{Output}: Model's parameters $\theta $
    \begin{algorithmic}[1] 
    \State Randomly initialize $\theta$.
    
    \For{$e=0$ to $E_{warmup}$} \Comment{standard ERM} 
        \State $\text{Train}(\tilde{\mathcal{D}},\theta)$ 
    \EndFor
    
    \For{$e=1$ to $E$} 
        \For{$i=0$ to $N$}
            \State obtain $\mathcal{N}_k(i)$ based the learned representation $\bs z_i=g(\bs x_i)$
            \State obtain every sample's LLC score $\mathrm{LLC}_i=\sum_{j \in \mathcal{N}_k(i)} \mathds{1} [\tilde{\bs y}_i=\tilde{\bs y}_j]$ 
        \EndFor
        \State fit $\mathrm{GMM}$ on $\{\mathrm{LLC}_i\}_{i=0}^{N}$ and obtain per-sample confidence $\mathcal{W}=\{w_i\}_{i=0}^{N}$
        
        \State  Robust training for one epoch: $\theta=\text{RobustTrain}(\tilde{\mathcal{D}},\mathcal{W},\theta)$  following Alg. \ref{RobustTrain}
    \EndFor

    \end{algorithmic} 
\end{algorithm}
            
\begin{algorithm}
    \caption{RobustTrain} 
    \label{RobustTrain}
    \textbf{Input}: Noisy dataset $\tilde{\mathcal{D}}$, label confidence $\mathcal{W}$, batch size $B$
    \textbf{Output}: Model's parameters $\theta $
    \begin{algorithmic}[1] 
            \For{$n=0,1, ..., \frac{|\tilde{\mathcal{D}}|}{B}$}
                \State Randomly draw a mini-batch $\{(\bs x_i,\tilde{\bs y}_i,w_i)\}_{i=1}^B$ from $\tilde{\mathcal{D}}$ and $\mathcal{W}$
                \For{$i=0$ to $B$}
                    \State $\hat{\bs y}_i=\text{p}(\bs y\mid \alpha(\bs x_i);\theta)$ 
                    \Comment{pseudo-label}
                    \State $\bs y^*_i=w_i\tilde{\bs y}_i+(1-w_i)\hat{\bs y}_i$ \Comment{refurbish label}
                \EndFor
                \State $\mathcal{L}=\{\text{H}(\bs y_i^*, \text{p}(\bs y\mid \mathcal{A}(\bs x_i);\theta))\}_{i=1}^{B}$ 
                \State $\mathcal{L}'=\mathop{\arg \max}\limits_{\mathcal{L}' \subset \mathcal{L}} \sum_{\ell_i \in \mathcal{L}'} \ell_i \quad \mathrm{s.t.} \, |\mathcal{L}'|=|\mathcal{L}|*\tau\%$
                \State update $\theta$ according to the gradient $\nabla \frac{1}{|\mathcal{L}'|}\sum_{\ell_i \in \mathcal{L}'} \ell_i$
            \EndFor
    \end{algorithmic} 
\end{algorithm}

\subsection{Robust training}
\noindent
\textbf{Refurbish-DRO}
We aim to utilize the estimated label confidence to improve robustness against both label noise and subpopulation imbalance.
To this end, we proposed the refurbish-DRO paradigm.
Specifically, we first generate refurbished labels on top of it to supervise training \cite{reed2014training}, which is a convex combination between noisy labels and pseudo-labels:
\begin{equation}
    \label{eq_refur}
    \begin{aligned}
        \hat{\bs y}_i&=\text{p}(\bs y \mid \bs x_i;\theta)\\
        \bs y^*_i&=w_i\tilde{\bs y}_i+(1-w_i)\hat{\bs y}_i 
    \end{aligned}
\end{equation}
where $w_i$ is the estimated label confidence of $i$-th sample, pseudo-label $\hat{\bs y}_i$ are from model's predictions, and the refurbished label $\bs y^*_i$, replacing the original noisy label $\tilde{\bs y}_i$, acts as the corrected supervision signal, $\theta$ represents the model parameters.
The refurbishment can be seen as a balance between given supervision (noisy labels) and self-supervision (pseudo-labels), where the label confidence is the weight to adaptively adjust the contribution of each component.

Then, we attempt to further ensure the model's fairness for tail subpopulations or minority samples (i.e., \textit{robust accuracy}).
To solve this problem, previous studies \cite{DBLP:conf/icml/HuNSS18,DBLP:journals/corr/abs-1911-08731} seek to minimize the worst-case empirical risk, 
\begin{equation}
\label{robust_loss}
    \mathop{\arg \min}\limits_{\theta}\left\{ \mathop{\max}\limits_{k\in {1,\ldots,K}} \frac{1}{|\mathcal{D}_k|} \sum_{(x_i,y_i) \in \mathcal{D}_k}\ell_i \right\}
\end{equation}
where $\ell_i$ can be any surrogate loss function.

However, Eq. (\ref{robust_loss}) needs group annotations and cannot cope with noisy labels.
Thus, we attempt to design a robust training method that does not require group annotations and is also applicable in the coexistence of label noise.
It has been found that deep models tend to fit head groups with commonly shared patterns (e.g., water background for waterbirds) instead of tail groups with infrequent patterns (e.g., land background for waterbirds) \cite{DBLP:journals/corr/abs-1911-08731,DBLP:conf/icml/LiuHCRKSLF21}.
Therefore, the tail subpopulations can be heuristically identified by the discrepancy between predicted probabilities and the ground-truth labels \cite{DBLP:conf/icml/LiuHCRKSLF21,DBLP:journals/corr/abs-2007-02561,DBLP:conf/icml/Zhai0KR21}.
Those with bigger losses (between the ground-truth label) are more likely to come from more challenging groups.
To approximate Eq. (\ref{robust_loss}), a simple but effective modification of the original loss is adopted.
Specifically, we rank the samples in every mini-batch based on the loss between the model's predicted distribution and the refurbished label distribution and only select the top-$\tau$\% for training:
\begin{equation}
    \mathop{\arg \min}\limits_{\theta} \left\{\mathop{\max}\limits_{\mathcal{L}' \subset \mathcal{L}} \sum_{\ell_i \in \mathcal{L}'} \ell_i \right\} \quad \mathrm{s.t.} \, |\mathcal{L}'|=|\mathcal{L}|*\tau\%
\end{equation}
where $\ell_i$ is the loss of $i$-th sample, $\mathcal{L}$ is a set of loss values of samples in the current batch.
We instead use the obtained refurbished label $\bs y^*_i$ in Eq. (\ref{eq_refur}):
\begin{equation}
\label{eq_ce}
\begin{aligned}
    \ell_i &= \text{H}(\bs y^*_i, \text{p}(\bs y\mid \bs x_i;\theta))\\
\end{aligned}
\end{equation}
where $\mathrm{H}(\cdot)$ is the cross-entropy function.
In this way, we avoid using the given noisy labels to calculate the loss and obtain the uncertainty set $\mathcal{L}'$ \cite{DBLP:conf/icml/LiuHCRKSLF21}, which would certainly fail due to noisy samples' interference. 
Our final learning target, therefore, focuses more on the difficult-to-learn samples while preventing the model from potentially overfitting the head subpopulation.

\noindent
\textbf{Weak-strong augmentation schema}
To generate high-quality pseudo-labels and improve generalization, we employ the recently weak-strong data augmentation schema \cite{sohn2020fixmatch}.
Specifically, The model's predictions on the weakly augmented images $\alpha(\cdot)$ is used to generate pseudo-labels, i.e., we change the first equation in (\ref{eq_refur}) to:
\begin{equation}
    \hat{\bs y}_i=\text{p}(\bs y\mid \alpha(\bs x_i);\theta)
\end{equation}
The model's predictions on the strongly augmented images $\mathcal{A}(\cdot)$ (e.g., applied with multiple geometric and color transformations) are enforced to be consistent with the pseudo-labels, i.e., we change the loss in equation (\ref{eq_ce}) to:
\begin{equation}
    \text{H}(\bs y_i^*, \text{p}(\bs y\mid \mathcal{A}(\bs x_i);\theta))
\end{equation}
The weak-strong augmentation schema in our LNL method can be seen as a special variant of consistency regularization or self-training method.
We set the weak augmentation $\alpha$ as a random crop followed by a random horizontal flip.
For the strong augmentation $\mathcal{A}$, we use the FixMatch version of RandAugment \cite{cubuk2020randaugment}.
More details are given in the supplementary material.

\noindent
\textbf{Co-training} Following previous works \cite{han2018co,li2020dividemix}, two models with the same structure but different random initializations are simultaneously trained. 
We use the confidence produced by one model to guide the training of its peer, i.e., the estimated label confidence from one model is used by another model and \textit{vice versa}.
Furthermore, we also use the co-teaching schema as in \cite{li2020dividemix}, i.e., pseudo-labels are generated from the ensemble (average) predictions of two models.

\section{Experiments}
\label{sec_experiments}

\begin{table*}
\centering
\begin{tabular}{l|cc|cc|cc|cc}
\toprule
Dataset        & \multicolumn{8}{c}{CelebA}           \\ \midrule 
Noise rate     & \multicolumn{2}{c|}{15\% }                & \multicolumn{2}{c|}{20\%}             & \multicolumn{2}{c|}{25\% }           & \multicolumn{2}{c}{30\%}           \\ \midrule 
Method/Group   & Avg. & Worst                              & Avg. & Worst                         & Avg. & Worst                         & Avg. & Worst                       \\ \midrule
ERM            & \textbf{95.60} & 34.44                    & \textbf{95.50} & 28.88               & \textbf{95.53} & 30.00	         & 95.20	& 20.00               \\    
DivideMix$^*$ & 95.20	  & 28.33                          &  95.22	& 28.88                   & 95.14 & 31.11                        & 95.05 & 20.55                      \\  
Ours           & 95.12 & \textbf{36.66}                    & 95.08 & \textbf{34.44}               &  95.12 & \textbf{41.66}              & \textbf{95.24}& \textbf{28.33}     \\  \bottomrule \toprule
Dataset        & \multicolumn{8}{c}{Waterbirds}      \\ \midrule 
Noise rate     & \multicolumn{2}{c|}{25\% }           & \multicolumn{2}{c|}{30\%}             & \multicolumn{2}{c|}{35\%}            & \multicolumn{2}{c}{40\% }             \\ \midrule 
Method/Group   & Avg. & Worst                         & Avg. & Worst                         & Avg. & Worst                         & Avg. & Worst                          \\ \midrule
ERM            & 83.15	& 48.60                       & 78.65 & 52.96                        & 79.53	& 64.17	                    & 73.66	& 48.29                     \\ 
DivideMix$^*$ & 85.92	& 55.61	                      & 79.32	& 54.67                    & 74.37	& 47.82                     & \multicolumn{2}{c}{Not converged}     \\ 
Ours           & \textbf{88.40} & \textbf{70.56}      & \textbf{86.09} & \textbf{69.31}      &  \textbf{84.93} & \textbf{71.88}     & \textbf{79.77}	& \textbf{59.97}    \\  \bottomrule 
\end{tabular} 
\caption{Comparison over the corrupted Waterbirds and CelebA dataset. $^*$We use the same label confidence estimation method as DivideMix. For a fair comparison, we replace its other components and training schemas with ours.}
\label{table_subpopulation}
\end{table*}

We conduct experiments on different settings:
1). datasets with different levels of noisy labels and explicit subpopulation size imbalance.
2). datasets with both synthetic and real-world noisy labels (standard LNL benchmarks) that inevitably exist implicit subpopulation recognition difficulty or size imbalance.
All comparisons are made under the same model architecture, and the details are in the supplementary material.

\subsection{Comparison on corrupted datasets with explicit imbalanced subpopulations}
\label{sec_waterbird}

We evaluate our proposed method in the new setting of the coexistence of noisy labels and imbalanced subpopulation on the corrupted Waterbirds dataset \cite{DBLP:conf/iclr/SagawaKHL20} and corrupted CelebA dataset \cite{DBLP:conf/iccv/LiuLWT15}.
We use the Waterbirds dataset from \cite{DBLP:conf/iclr/SagawaKHL20}, which has 4,795 training samples consisting of waterbird and landbird images.
95\% of waterbirds are placed against water backgrounds, and the remaining 5\% of waterbirds are placed against land backgrounds.
Similarly, 95\% of landbirds are placed against land backgrounds, and the remaining 5\% of landbirds are placed against water backgrounds.
Then we randomly flip the labels of 25\%/30\%/35\%/40\% of samples to obtain the corrupted dataset.
CelebA's training set includes 71,629 females with dark hair, 66,874 males with dark hair, 22,880 females with blond hair, and only 1,387 males with blond hair.
The prediction task is to tell the color of the sample's hair.
Therefore, blond-haired men can be seen as the tail subpopulation, while others can be seen as head subpopulations.
It's also worth noting that the imbalance ratio of CelebA is bigger than that of Waterbirds.
Therefore, we apply lighter noise corruption, i.e., labels of 15\%/20\%/25\%/30\% of samples are corrupted to form the corrupted CelebA dataset.
These two datasets both have provided extra validation and test sets for hyper-parameter tuning and testing.
We save the best-performed model on the validation set and report its test accuracy.
To evaluate the model's performance on the overall population and tail subpopulation, we report the average test accuracy and the accuracy of the subpopulation that the model performs worst on, respectively. 

For comparison, we include two methods: 1). ERM, i.e., using no extra robustness training method to optimize the model. 
2). improved DivideMix. 
It uses DivideMix's label confidence estimation method but has the same training schema as ours to ensure a fair comparison.  
Backbone models of two baselines are also the same as our method.

In Table \ref{table_subpopulation}, we compare the proposed method with the improved DivideMix and the ERM model on two datasets.
The ERM model learns the spurious correlations, e.g., water background and waterbirds, which leads to worse performance on some tail subpopulations (i.e., worst-case accuracy) than that on average.
The improved DivideMix with the previous classification-loss-based confidence estimation also achieves low worst-case accuracy, sometimes even worse than the ERM baseline. 
This is because it would misclassify clean subpopulations into noise and exclude them from the training objective, aggravating the subpopulation imbalance problem.
Instead, our robust training method better combats spurious correlations on the corrupted Waterbirds dataset, yielding much better generalization ability on the subpopulations (worst-case accuracy improved by 7.7\%-14.95\%).
The model's average accuracy on the overall population also improved by 2.5\%-6.8\%. 
We provide more results on different imbalance ratios in the supplementary material.
On CelebA, our method also consistently achieves the best worst-case accuracy (improve the baselines by 2.2\%-10.5\%) while keeping the average accuracy greater than 95\%.

We also notice that the increase in noise rate does not necessarily lead to the decrease of worst-case performance in Table \ref{table_subpopulation}. 
We speculate that uniformly distributed noise may act as a regularization that encourages the model to make more diverse predictions.

\begin{table*}
    \centering
    \begin{tabular}{lr|c|c|c|c|c||c|c|c|c}
        \toprule
		Dataset               &      &\multicolumn{5}{c||}{CIFAR-10}& \multicolumn{4}{c}{CIFAR-100}\\\midrule
		Noise type &      &\multicolumn{4}{c|}{Sym.}& \multicolumn{1}{c||}{Asym.} & \multicolumn{4}{c}{Sym.}\\\midrule
		\multicolumn{2}{l|}{Method/Noise ratio}        & 20\% & 50\% & 80\% & 90\% & 40\% & 20\% & 50\% & 80\% &  90\% \\ \midrule
		\multirow{2}{*}{Co-teaching+}          & Best  & 89.5 & 85.7 & 67.4 & 47.9 &  -   & 65.6 & 51.8 & 27.9 & 13.7 \\
		                                       & Last  & 88.2 & 84.1 & 45.5 & 30.1 &  -   & 64.1 & 45.3 & 15.5 &  8.8 \\ \midrule
		\multirow{2}{*}{Meta-Learning}         & Best  & 92.9 & 89.3 & 77.4 & 58.7 & 89.2 & 68.5 & 59.2 & 42.4 & 19.5 \\
		                                       & Last  & 92.0 & 88.8 & 76.1 & 58.3 & 88.6 & 67.7 & 58.0 & 40.1 & 14.3 \\\midrule
		\multirow{2}{*}{M-correction}          & Best  & 94.0 & 92.0 & 86.8 & 69.1 & 87.4 & 73.9 & 66.1 & 48.2 & 24.3 \\
		                                       & Last  & 93.8 & 91.9 & 86.6 & 68.7 & 86.3 & 73.4 & 65.4 & 47.6 & 20.5 \\\midrule
		\multirow{2}{*}{DivideMix}             & Best  & \textbf{96.1} & 94.6 & 93.2 & 76.0 & 93.4 & 77.3 & 74.6 & 60.2 & 31.5 \\
				                               & Last  & 95.7 & 94.4 & 92.9 & 75.4 & 92.1 & 76.9 & 74.2 & 59.6 & 31.0 \\\midrule
		\multirow{2}{*}{Ours}            & Best & \textbf{96.1} & \textbf{95.9} & \textbf{95.9} & \textbf{94.8} & \textbf{94.7} & \textbf{79.2} & \textbf{75.5} & \textbf{67.7} &  \textbf{51.0} \\
				                         & Last & \textbf{96.0} & \textbf{95.8} & \textbf{95.8} & \textbf{94.6} & \textbf{94.1} & \textbf{78.1} & \textbf{74.5} & \textbf{65.3} & \textbf{49.2} \\
        \bottomrule
    \end{tabular}
    \caption{Comparison with state-of-the-art LNL methods on corrupted CIFAR. Sym. and Asym. are symmetric and asymmetric for short, respectively. The best results are indicated in bold. Other methods' results are from \cite{li2020dividemix}. } 
    \label{table_CIFAR}
\end{table*}

\begin{figure*}
    \centering
    \includegraphics[width=\textwidth]{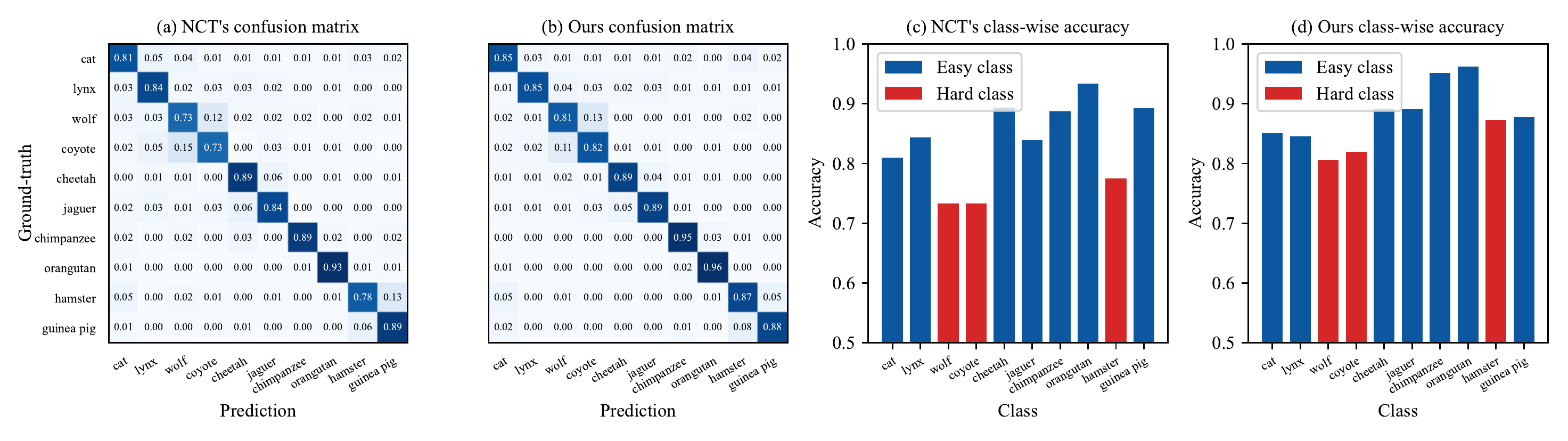} 
    \caption{The confusion matrix and class-wise accuracy of previous SOTA NCT and our method on ANIMAL-10N. 
    Other than average accuracy, our method achieves even more significant improvement in some difficult-to-learn classes. We define a class as hard when NCT's accuracy is lower than 80\%.
    We reproduce the results of NCT using their official GitHub code.} 
    \label{animal_confusion}
\end{figure*}

\subsection{Comparison with state-of-the-art methods on standard LNL benchmarks}
We further benchmark the proposed method on common LNL experimental settings using CIFAR \cite{krizhevsky2009learning} with different levels and types of synthetic noises, as well as the real-world noisy dataset Mini-WebVision  \cite{li2017webvision} and ANIMAL-10N \cite{song2019selfie}. 
We use ANIMAL-10N as an example of a real-world noisy dataset with implicit subpopulation imbalance.

\subsubsection{CIFAR-10, CIFAR-100 with synthetic label noise}
\label{sec:cifar}

Performance on randomly generated uniform (symmetric) noise can generally indicate the effectiveness of robust training methods.
Thus, we compare our method with 
Co-teaching+ \cite{yu2019does}, Meta-learning \cite{li2019learning}, DivideMix \cite{li2020dividemix} on uniformly corrupted CIFAR. 
Besides, following previous work \cite{li2020dividemix}, we also experiment with CIFAR-10 corrupted by a pre-defined transition matrix.
The corruption mimics the real-world noise in which samples are liked to be labeled into similar classes, e.g., deer to horse.
We report the performance of the last epoch and the best epoch following previous methods.

Table \ref{table_CIFAR} shows that our method consistently achieves competitive performance against previous methods under various noise rates and synthetic noise types.

\subsubsection{Mini-WebVision and ANIMAL-10N with real-world label noise}
The label noise and distribution shift could be more complex in the wild \cite{DBLP:journals/corr/abs-2011-04406}.
Therefore, it is necessary to conduct experiments on real-world noisy datasets.
WebVision \cite{xiao2015learning} includes large-scale web images. 
We follow \cite{chen2019understanding}, using the first 50 classes.
Besides, the validation set of ImageNet ILSVRC12 \cite{deng2009imagenet} is used.
ANIMAL-10N \cite{song2019selfie} includes 10 animals, where 5 pairs of them are alike, such as cat vs. lynx and jaguar vs. cheetah.

For comparison on Mini-WebVision, results of Co-teaching \cite{han2018co}, DivideMix \cite{li2020dividemix}, NGC \cite{wu2021ngc} are reported. 
Our method outperforms previous methods by 1.56\% top-1 accuracy on the test set, as shown in Table \ref{table_webvision}.
The results validate the effectiveness of our approach on real-world noise.

\begin{table}
\centering
\begin{tabular}{l|cc|cc}
    \toprule
    \multirow{2}{*}{Method} & \multicolumn{2}{c|}{Mini-WebVision} & \multicolumn{2}{c}{ILSVRC12} \\ \cmidrule(r){2-3}\cmidrule(l){4-5}
                                & top-1  & top-5  & top-1  & top-5  \\ \midrule
        Co-teaching             & 63.58 & 85.20 & 61.48 & 84.70 \\
        DivideMix               & 77.32 & 91.64 & 75.20 & 90.84 \\ 
        NGC                     & 79.16 & 91.84 & 74.44 & 91.04 \\
        Ours                      & \textbf{80.72} & \textbf{93.19} & \textbf{76.76} & \textbf{93.76} \\ \bottomrule
    \end{tabular}
\caption{Comparison with other methods on Mini-WebVision. The results of other methods are from \cite{li2020dividemix,wu2021ngc}.}
\label{table_webvision}
\end{table}

We choose NCT \cite{DBLP:conf/cvpr/ChenSHS21}, the previous best method on ANIMAL-10N, for the comparison in Fig. \ref{animal_confusion} (comparison with more methods is in the supplementary material).
ANIMAL-10N, as a real-world class-balanced dataset, has different learning difficulties across classes.
Wolf, coyote, and hamster are harder to recognize for NCT, as shown in Fig. \ref{animal_confusion} (a) and (c).
The results support our argument that real-world datasets could have implicit ``imbalanced'' subpopulations varying in learning difficulties.
Our average accuracy is 4.1\% higher than NCT (88.2\% vs. 84.1\%), while the improvement mainly comes from the more difficult classes (i.e., those NCT performs worse on), e.g., on wolf is 80.6\% vs. 73.4\% and on coyote is 82.0\% vs. 73.3\%.
Our model achieves better worst-case (in terms of class) performance, which verifies that our method successfully addresses such imbalance, improving generalization ability.

\subsubsection{Ablation study}
\label{sec:ablation}
We conduct the ablation study to verify the effectiveness of the two main components, i.e., LLC and refurbish-DRO.
We replace LLC with classification-loss-based label confidence estimation or remove the DRO strategy and compare the performances of these variants.
The performances of four combinations are illustrated in Table \ref{table_ab}.
The proposed two components both bring remarkable improvement upon the baselines.
Adding the refurbish-DRO strategy on both classification-loss-based LNL (second row vs. first row) and feature-based LNL (fourth row vs. third row) can improve worst-case performance.
The combination of LLC and refurbish-DRO strategy achieves consistent best results under various noise rates.

\begin{table}
    \centering
    \resizebox{\columnwidth}{!}{%
    \begin{tabular}{cc|cc|cc}
    \toprule
         \multicolumn{2}{c|}{Component/Group} & \multicolumn{2}{c|}{25\% noise rate}      & \multicolumn{2}{c}{30\% noise rate} \\\cmidrule(lr){1-2} \cmidrule(lr){3-4}\cmidrule(lr){5-6}
         LLC          &  Refurbish-DRO & Avg. & Worst & Avg. & Worst \\ \midrule 
                      &            & 85.92	& 55.61	& 79.32	& 54.67 \\
                      &\checkmark  & 84.90     & 59.81 & 79.58 & 55.29 \\ 
        \checkmark    &            & 87.63 & 67.60 & 82.14 & 64.39 \\ 
        \checkmark    &\checkmark  &\textbf{88.40} & \textbf{70.56} & \textbf{86.06} & \textbf{69.31} \\  \bottomrule \toprule
        \multicolumn{2}{c|}{Component/Group} & \multicolumn{2}{c|}{35\% noise rate}      & \multicolumn{2}{c}{40\% noise rate} \\\cmidrule(lr){1-2} \cmidrule(lr){3-4}\cmidrule(lr){5-6}
         LLC          &  Refurbish-DRO & Avg. & Worst & Avg. & Worst \\ \midrule 
                      &            & 73.95	& 26.79 & \multicolumn{2}{c}{Not converged} \\ 
                      &\checkmark  & 74.54  & 49.22 & 79.39 & 49.06 \\ 
        \checkmark    &            & 82.50	& 65.90	& 74.78	& 51.09 \\
        \checkmark    & \checkmark &  \textbf{84.93} & \textbf{71.88} & \textbf{79.77}	& \textbf{59.97} \\  \bottomrule
    \end{tabular}
    }
    \caption{Ablation study on the corrupted Waterbirds dataset.}
    \label{table_ab}
\end{table}

\begin{figure}
    \centering
    \includegraphics[width=\columnwidth]{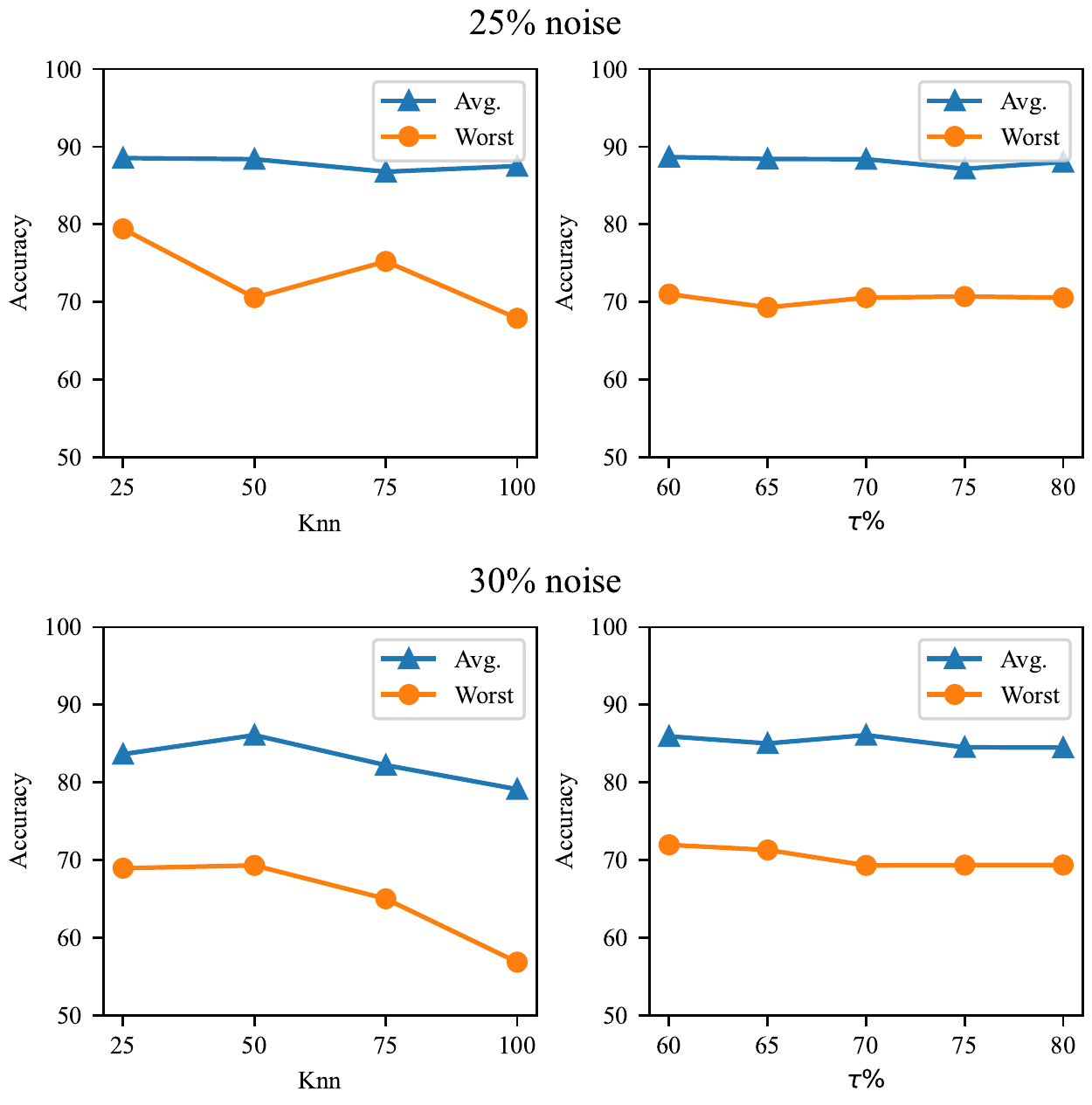} 
    \caption{
    Hyper-parameter sensitivity tests for the nearest neighbor numbers $k$ and threshold $\tau$ for refurbish-DRO on the corrupted Waterbirds dataset under 25\% and 30\% noise.} 
    \label{fig_senitivity}
\end{figure}

We also conduct sensitivity tests for two hyper-parameters, i.e., nearest neighbor number $k$ in KNN for the calculation of LLC and the threshold $\tau$ for refurbish-DRO. The results are given in Fig. \ref{fig_senitivity}, where we found that our proposed method can generally achieve robust high performance with different choices of $k$ and $\tau$.

\section{Conclusion}
This paper studies a realistic but challenging problem: learning with noisy labels over imbalanced subpopulations.
Plenty of works concentrating on learning with noisy labels or distribution shifts (imbalanced subpopulations) rarely address the contradiction in the coexistence of label noise and subpopulation shifts. How to deal with those samples that are perceptually inconsistent with the learned model but could be either informative or noisy is still an open question.
As a pioneer exploration, our method first resorts to label confidence estimation based on label consistency in feature space, alleviating the misidentification in conventional classification-loss-based LNL methods.
Using the estimated label confidence, we propose the refurbish-DRO paradigm, making the model less susceptible to spurious correlations.
Extensive experiments verify the effectiveness of our method.
Our work has the potential to motivate researchers to explore more realistic problems to facilitate the robust deployment of deep models in the wild.

{\small
\bibliographystyle{ieee_fullname}
\bibliography{egbib}
}
\clearpage

\end{document}